\documentclass[journal]{IEEEtran}
\usepackage{preambles}

\title{Machine Vision-Enabled Sports Performance Analysis}

\author{Timilehin B. Aderinola$^{*\xi}$, Hananeh Younesian$^\xi$, Cathy Goulding, Darragh Whelan,
        \\Brian Caulfield, Georgiana Ifrim
    \thanks{$^\xi$ These authors contributed equally to this work.}
    \thanks{This work was funded by Enterprise Ireland under grant number IP20210963E and Science Foundation Ireland through the Insight Centre for Data Analytics (12/RC/2289\_P2).}
    \thanks{T.B. Aderinola, H. Younesian, B. Caulfield, and G. Ifrim are with the Insight SFI Centre for Data Analytics, University College Dublin, Ireland. D. Whelan and C. Goulding are with Output Sports, NovaUCD, University College Dublin, Ireland. $^{*}$Corresponding author. Email: \href{mailto:timilehin.aderinola@ucd.ie}{timilehin.aderinola@ucd.ie}.}
}

\begin{document}
\maketitle

\begin{abstract}
\textbf{\textit{Goal:}} This study investigates the feasibility of monocular 2D markerless motion capture (MMC) using a single smartphone to measure jump height, velocity, flight time, contact time, and range of motion (ROM) during motor tasks. \textbf{\textit{Methods:}} Sixteen healthy adults performed three repetitions of selected tests while their body movements were recorded using force plates, optical motion capture (OMC), and a smartphone camera. MMC was then performed on the smartphone videos using OpenPose v1.7.0. \textbf{\textit{Results:}} MMC demonstrated excellent agreement with ground truth for jump height and velocity measurements. However, MMC's performance varied from poor to moderate for flight time, contact time, ROM, and angular velocity measurements. \textbf{\textit{Conclusions:}} These findings suggest that monocular 2D MMC may be a viable alternative to OMC or force plates for assessing sports performance during jumps and velocity-based tests. Additionally, MMC could provide valuable visual feedback for flight time, contact time, ROM, and angular velocity measurements.
\end{abstract}

\renewcommand\IEEEkeywordsname{Impact Statement}
\begin{IEEEkeywords}
Smartphone-based motion capture offers a viable and accessible tool for assessing sports performance with a simple setup that requires no camera calibration, particularly for jump height and velocity measurements. 
\end{IEEEkeywords}

\section{Introduction}\label{sec:introduction}

Objective performance assessment is essential in sports for strength and conditioning, injury risk screening, and rehabilitation~\cite{oreilly2018wearable}. For example, athletes' lower-limb explosive power can be assessed based on a countermovement jump~\cite{cmj_review}; velocity-based tests (VBT) such as overhead press and back squat enable objective recommendation of appropriate resistance training for athletes \cite{vbt}; and the range of motion measured during internal and external hip rotation can be used for injury risk screening in baseball players \cite{bullock2020hip}. Traditionally, assessment is performed by trained analysts. However, such assessments are subjective due to differences in exercise capture technique, personal bias, and method of analysis \cite{whelan_delahunt_2019}. 

More recently, motion capture systems have been proposed for more objective assessment and analysis of performance during motor tasks. These motion capture systems can be broadly classified as marker-based or markerless. An example of marker-based motion capture is optical motion capture (OMC), which involves the careful placement of light-emitting diodes (LEDs) or reflective markers for keypoint tracking. However, not only are marker-based devices such as OMCs expensive, they require technical expertise and a long setup time. Moreover, physical markers may change position during tasks such as jumps, which may reduce reliability in performance assessment. Recent advances in computer vision have enabled the automatic tracking of human motion via markerless motion capture (MMC), which enables objective sports performance analysis \cite{cust_sweeting_ball_robertson_2018}, such that commercial MMC systems \cite{theia_paper, captury_2019} are being deployed in practice.

Current MMC methods are either monocular or multi-camera. Monocular MMC relies on 2D images captured from a single camera \cite{under_water_running,outside_the_lab,aderinola2023quantifying}, while multi-camera MMC requires careful calibration of multiple cameras and 3D pose reconstruction from several 2D images \cite{nakano2020evaluation, corazza_mmc}. Although 3D multi-camera techniques are highly accurate, obtaining 2D recordings from single smartphone cameras is easier, more convenient, and cheaper. However, existing approaches for quantifying body kinematics using 2D monocular MMC have certain limitations such as being based on deep learning approaches \cite{under_water_running,outside_the_lab}. While deep learning can achieve very good results, its generalisation ability will depend on the size and diversity of the data. However, the collection of sufficiently large representative data is difficult. In \cite{my_jump}, an alternative quantitative approach is taken, but the proposed system requires manual selection of exercise duration and distance calibration. 

Distance calibration is an important challenge in 2D monocular MMC because most 2D cameras have no depth information. Therefore, some recent studies \cite{gravity_ref, aderinola2023quantifying} have proposed ways of automatic distance calibration by performing pixel-to-metric conversion for each video using gravity as a real-world reference. However, these and similar studies focused on single tasks such as countermovement jumps and underwater running \cite{under_water_running}. Here, we further evaluate 2D MMC on a much wider range of motor tasks (Table~\ref{tab:tasks}) captured with a single smartphone camera. More particularly, the objectives of this work are:
\begin{enumerate}
    \item  To evaluate the performance of MMC in quantifying jump heights using force plates as ground truth.
    \item To evaluate the performance of MMC in quantifying velocity using optical motion capture as ground truth.
    \item To evaluate the performance of MMC in quantifying temporal metrics such as flight time and contact time using the force plate as ground truth.
    \item To evaluate the performance of MMC in quantifying angular metrics using optical motion capture as ground truth.
\end{enumerate}

\begin{figure}[ht]
  \centering
  \includegraphics[width=1\linewidth]{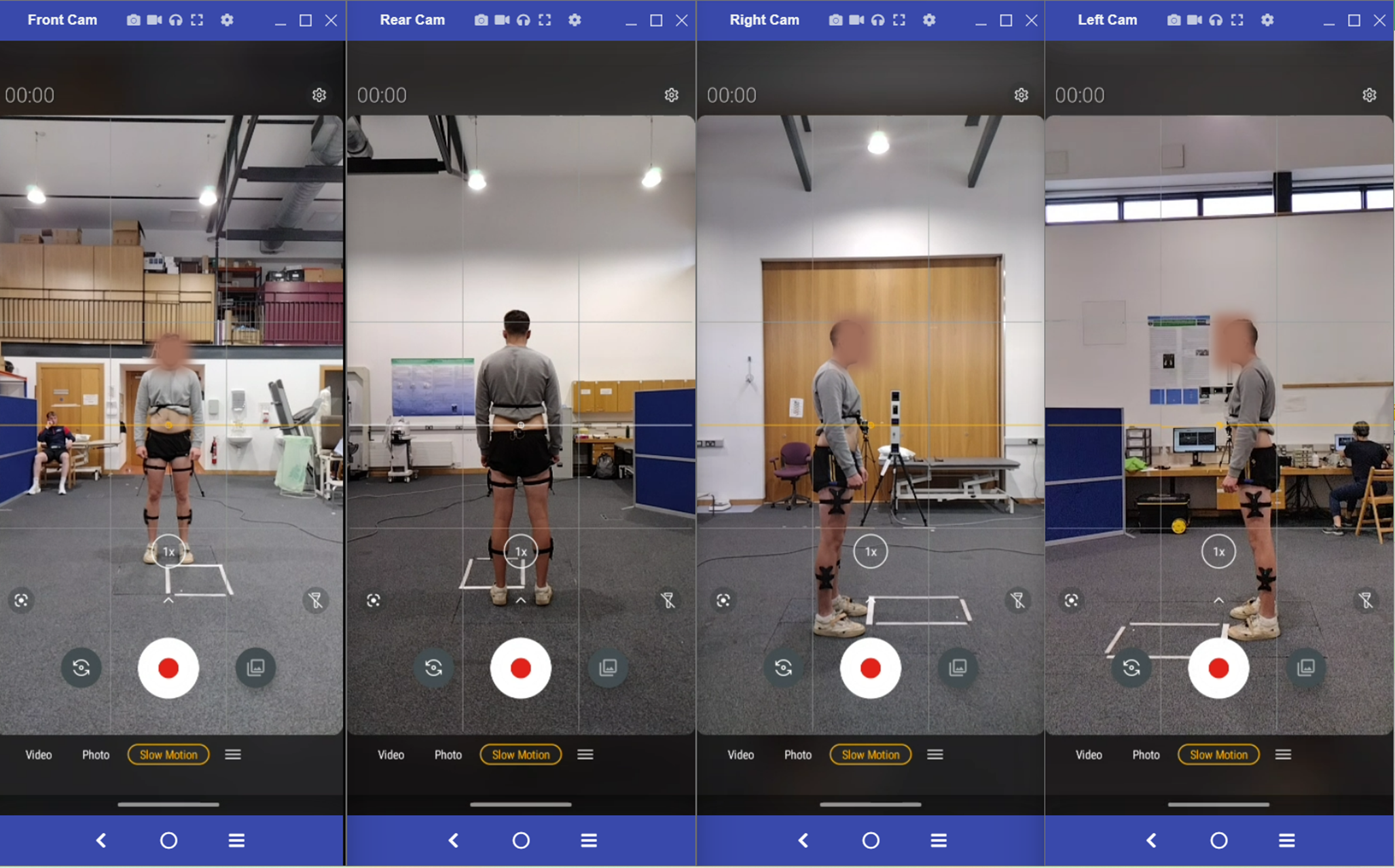}
  \caption{\textbf{Camera Setup}. All the tasks were recorded using cameras on the front, rear, left, and right sides of the participants. However, we only used a single view for each task.}
  \label{fig:camera-setup}
\end{figure}

To accommodate the multi-view recording requirements, we employed four smartphones positioned to capture front, rear, left, and right views of the participants (Fig. \ref{fig:camera-setup}). However, prioritizing ease of deployment and usability, each task analysis utilizes the recording from a single smartphone without distance calibration. 

\textbf{Our main contributions are:}
\begin{enumerate}
    \item We propose a simple setup that is both affordable and realistic in settings where performance is assessed outside the lab, for example, at home.
    \item We implement and compare three methods of converting image pixel distances to real-world distances and their impact on performance. This removes the need for manual distance calibration.
    \item We analyse the agreement of MMC with OMC or Force Plates as ground truth in quantifying jump height, velocity, temporal metrics, and angular metrics across a wide range of motor tasks.
\end{enumerate}

To facilitate reproducibility, the code and data necessary to reproduce our analysis are available on GitHub\footnote{\href{https://github.com/mlgig/mvespa}{https://github.com/mlgig/mvespa}}.

\begin{table*}
  \centering
  \caption{Description of Tasks}
  \label{tab:tasks}
  \begin{tabular}{lllc}
    \toprule
    \textbf{Task} & \textbf{Description} & \textbf{Target Metrics}  &\textbf{Camera View}\\
    \midrule
    BSQ & Back squat & Barbell velocity  &Front\\
    OHP & Overhead press & Barbell velocity  &Front\\
    CMJBL & Countermovement jump (bilateral) & Jump height  &Front\\
    CMJUL & Countermovement jump (unilateral) & Jump height  &Front\\
    DJBL & Drop jump (bilateral) & Jump height, flight time, contact time  &Front\\
    DJUL & Drop jump (unilateral) & Jump height, flight time, contact time  &Front\\
     RJT & 10-5 Test (repeated jump test) & Jump height, flight time, contact time  &Front\\
    NDC & Nordic curl & Angular velocity, Range of motion  &Right\\
    SLS & Single leg squat & Angular velocity, Range of motion  &Right\\
    HER & Seated hip external rotation & Range of motion  &Front\\
    HIR & Seated hip internal rotation & Range of motion  &Front\\
    SLR & Straight leg raise & Range of motion  &Right\\
    \bottomrule
  \end{tabular}
\end{table*}


\section{Methods}\label{sec:methods}

\subsection*{Data Collection}
Sixteen healthy adult volunteers (Table 1, Supplementary Information) participated in this study. We determined the dominant foot of each participant based on the foot with which they kicked a ball \cite{kick_a_ball}. Each participant signed the informed consent form approved by the Human Research Ethics Committee of University College Dublin with Research Ethics Reference Number LS-C-22-117-Younesian-Caulfield. After a five-minute warm-up, each participant performed three repetitions of the twelve target tasks (Table~\ref{tab:tasks}) while simultaneous capture of the tasks was carried out using optical motion capture and markerless motion capture.

\subsection{Apparatus}
\label{sec:apparatus}

\subsubsection{Optical Motion Capture}
\label{sec:omc}
Optical motion capture was performed using four synchronised codamotion (CODA, Charnwood Dynamics, UK) 3D cameras sampling at 100 Hz. For the upper-limb tasks, two light-emitting diode (LED) markers were placed on the barbell - one marker at either end of the barbell to track the motion of the barbell. For the lower-limb tasks, four LED marker clusters were placed on the left and right lateral sides of the thigh and shank (Fig. 8, Supplementary Information). Each marker cluster consists of four markers. In addition, six LED markers were placed on the superior iliac crest (anterior and posterior), and greater trochanter (left and right). Three LED markers were attached to the lateral side of the calcaneus and on the first and fifth metatarsals of the dominant foot. The knees and ankles were tracked with virtual markers using the cluster markers as reference.

\subsubsection{Markerless Motion Capture}
\label{sec:mmc}
Markerless motion capture was performed using four unsynchronised Motorola G4 smartphone cameras recording at a resolution of 720p and a frame rate of 30 frames per second (fps). The cameras were placed on tripods positioned in front, to the left, right, and at the back of the participant (Fig. \ref{fig:camera-setup}). There were no strict requirements on the distance to the participant, but it was ensured that the cameras remained stationary and participants remained fully visible in the camera views.

We performed 2D HPE using OpenPose v1.7.0 \cite{openpose} to obtain key body points from the recorded videos on a Linux Ubuntu 18.08 server with 20GB RAM. All analyses were performed on a PC running Windows 11, with an Intel core i7 processor and 16GB RAM. Python 3.8 was used for data analysis and visualisations.

\subsection{Data Preprocessing}
\label{sec:preprocessing}
Given a test lasting $T$ seconds and $K$ markers, the output from OMC is a sequence of 3D coordinates \(\{(x_i^t,y_i^t,z_i^t ) | i=1,...,K;t=1,...,100T\}\) in \textit{millimeters}, where 100 is the frequency of motion capture. The output from MMC is a sequence \(\{(x_i^t,y_i^t,c_i^t )| i=1,...,K;t=1,...,30T\}\), where 30 is the frame rate, \((x_i^t,y_i^t)\) are the 2D coordinates in \textit{pixels}, and \(c_i^t \in [0,1]\) is the confidence score for part  $i$ in frame $t$. The data preprocessing steps performed include smoothing, synchronisation, and rescaling.

\subsubsection{Smoothing}
Motion time series obtained from human pose estimation (HPE) are often noisy. There are a number of reasons for this, which include the video quality, amount of background clutter, video viewpoint, and false positives~\cite{aderinola2023quantifying}. We performed smoothing with a second-order Savitzy-Golay (Savgol) filter~\cite{savitzky1964smoothing}, which preserves the extrema of the motion time series.

\subsubsection{Segmentation and Synchronisation}
\label{sec:sync}
 
\begin{figure}[ht]
  \centering
   \includegraphics[width=1\linewidth]{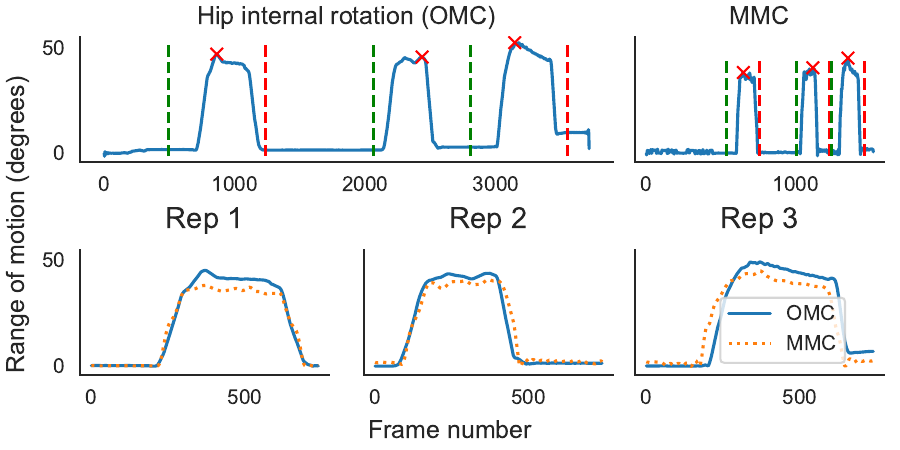}
   \caption{\textbf{Segmentation and Synchronisation}. Segmentation and synchronisation of repetitions illustrated with the seated hip internal rotation. Each repetition is identifiable by a maxima, indicated with \textcolor{red}{$\times$}, which represents the maximum rotation. The repetition windows are bounded with the dashed vertical lines.}
   \label{fig:seg_sync}
\end{figure}

For some tasks, such as the drop jump, segmentation and synchronisation were performed manually. However, for most tasks, segmentation was based on the maxima and frequency of the motion time series  (Fig. \ref{fig:seg_sync}).

Let $s_{x}$ be a maxima of a univariate motion time series $\{s_i|i=1,2,..., L\}$, we segmented repetition $r$ using $t_1$ seconds before and $t_2$ seconds after $s_{x}$:
\begin{equation}
    r = \{s_{x- ft_1}, ..., s_{x},..., s_{x+ ft_2}\},
\end{equation}
 where $f$ is the capture frequency, $x-ft_1 \geq 1$, and $x+ft_2 \leq L$. This method of segmentation ensures that corresponding segmented repetitions from each motion capture technique are of approximately equal duration in seconds. However, due to the difference in capture frequency, the repetitions from the capture devices were of different lengths. Hence, we upsampled the MMC time series to be of the same length as the OMC time series using Fast Fourier Transform resampling \cite{fourier_resample}, which resamples with minimal distortion.

\subsubsection{Rescaling}
\label{sec:rescaling}
Pose estimation models measure positions and distances in pixels (\emph{px}). To enable the comparison of distance measurements, we explored three references to perform pixel-to-metric conversion (PTM). These are gravity, (PTM$_g$) \cite{aderinola2023quantifying,gravity_ref}, subject's height (PTM$_h$), and an object of known length (PTM$_b$). The rescaling step was performed for each participant since they all stood at different distances from the camera.

\textbf{Gravity as a reference (PTM$_g$)}. This is a generalised method that takes advantage of universal gravity and requires a bilateral jump. It involves three main steps. First, the free-fall distance in metres $d_{m}$ travelled by the participant's centre of mass (COM) after $t$ seconds is determined according to Newton's equation of free-fall motion
\begin{equation}\label{eq:gravity1}
    d_{m}(t) = d_0 + v_0t + gt^2/2
\end{equation}
where $g$ is the universal acceleration due to gravity, and the initial distance $d_0$ and initial velocity $v_0$ are zero. Hence, we have
\begin{equation}\label{eq:gravity}
    d_{m}(t) = gt^2/2
\end{equation}
Next, the actual pixel free-fall distance in time $t$, $d_{px}$, is measured. From this, the PTM scale is determined as:
\begin{equation}\label{ptmg}
    PTM_g = d_m/d_{px}
\end{equation}

\textbf{Subject's height as a reference (PTM$_h$)}. This method reasonably assumes that any exercise application will collect users' demographic information such as height in metres. To use the subject’s height as a reference (PTM$_h$), we obtained the PTM scale as
\begin{equation}\label{ptmh}
    PTM_h = h_{m}/h_{px}
\end{equation}
where $h_m$ and $h_{px}$ represent the participant's height in metres and pixels respectively. This is a personalised method and is more convenient. However, since the anatomical limits of OpenPose are the nose and toe, we referred to anthropometric ratios \cite{casadei_kiel_2022} to estimate the full pixel heights of participants. In particular, we estimate the total body height of each participant by taking the shoulder-to-toe distance as $6/8$ of the total height of an adult.

\textbf{Object of known length as a reference (PTM$_b$)}. This is also a personalised method for tasks with objects of known length such as barbells. In this method, we obtained the PTM scale as
\begin{equation}\label{ptmb}
    PTM_b = l_{m}/l_{px}
\end{equation}
where $l_m$ and $l_{px}$ represent the length of the barbell in metres and pixels respectively. The length of the barbell was measured to be 1.125m. The pixel length of the barbell for each participant was measured using an on-screen pixel ruler scaled to the resolution of the videos.

\subsection{Quantifying Target Metrics}

\subsubsection{Jump Height}
\label{sec:cmj}
The vertical jump height was measured during countermovement jumps and drop jumps by subtracting the maximum displacement of the toe from its resting position. Drop jumps were performed from a 22 cm high platform (Fig. 1, Supplementary Information). In the side view, several unilateral jumps had to be discarded because the pose estimation model confuses the left and right limbs (Fig. \ref{fig:noisy}). Hence, better performance for countermovement and drop jumps could be achieved in the front view.

\subsubsection{Velocity}
\label{sec:vbt}

Two velocity-based tests, namely overhead press (OHP) and back squat (BSQ), were captured in the front view. To avoid fatigue and injury, a dummy barbell was used. The barbell's maximum velocity (m/s) and mean velocity (m/s) were measured using MMC and compared to the OMC as ground truth. The velocity was measured by differentiating the vertical motion of the barbell during the concentric phase of the task with respect to time (Fig. 2, Supplementary Information). The concentric phase is the phase during which the muscles rise against the weight of the barbell. For MMC, the wrist was used as a surrogate for the barbell since OpenPose tracks only human keypoints.

\subsubsection{Temporal Metrics}
\label{sec:time}

Flight time and contact time were measured during the 10-5 test, also known as the repeated jump test (RJT), and drop jumps (DJBL and DJUL). For RJT, the contact time and flight time were measured directly from the MMC time series. For the drop jumps, the contact time was obtained as the distance between the first two peaks of the toe velocity, while the flight time was obtained as the distance between the last two peaks (Fig. 3, Supplementary Information). 

\subsubsection{Angular Metrics}
\label{sec:angles}

Range of motion (ROM) was measured during hip internal rotation (HIR), hip external rotation (HER), Nordic Curl (ND), straight leg raise (SLR), and single leg squat (SLS). Mean angular velocity was measured during the Nordic Curl and Single Leg Stance. HIR and HER were performed in the seated position and captured from the front view with OMC as the ground truth. We measured the total ROM as the maximum angular displacement between the resting position and the final position of the ankle (Fig. 4, Supplementary Information).

In particular, we used the angular displacement at the knee as a proxy to determine the hip ROM. To do this, we took the tibia (the shin bone between the knee and ankle) as a vector in 2D space. Let $\vec{ka_0}$ be the tibia at the resting position, $\vec{ka_i}$ the tibia after internal rotation, and $\vec{ka_e}$, the tibia after external rotation. We computed the hip internal ROM ($\theta_i$) and external ROM ($\theta_e$) by finding the acute angle of intersection between $\vec{ka_0}$ and $\vec{ka_e}$, 
\begin{equation}\label{eq:angle}
    \theta_p = \tan^{-1}|(m_{ka_p} - m_{ka_0})/(1 + (m_{ka_p} \times m_{ka_0}))| 
\end{equation}
where $m_v$ is the slope of $v$, and $p \in \{i, e\}$ for internal and external rotation respectively. An example of a participant's hip internal and external rotation is shown in Fig. 4 of the Supplementary Information. 

During the Nordic curl, the femur and tibia were taken as vectors $\vec{f}$ and $\vec{t}$ respectively, and the angular displacement was measured as the angle $\theta$ between them:
\begin{equation}\label{eq:nordic-angle}
    \theta = \cos^{-1}({f}\cdot{t}/|f||t|) 
\end{equation}
The range of motion was measured as the maximum angular displacement, while the mean angular velocity was measured as the mean velocity from the start of the curl to the peak of the velocity (Fig. 5, Supplementary Information).

The straight leg raise range of motion was measured by taking the whole leg (hip to ankle) as a single vector and measuring its angular displacement relative to the resting position (Fig. 6, Supplementary Information). 

During the single leg squat (SLS), the participant raises a foot and performs a squat on the dominant foot. The angular displacement is measured at the knee of the dominant foot as the angle between the femur and tibia (Fig. 7, Supplementary Information). The SLS and SLR are possible only from the side view in the 2D plane. For several participants and repetitions, the pose estimation model confused the left and right limbs during SLR and SLS (Fig. 6b, 7b, Supplementary Information). Therefore, for the SLR, the data for 9 participants were discarded and only 7 participants were used for analysis.

\section{Results}\label{sec:results}

\begin{table*}
\caption{Performance Overview of 2D Markerless Motion Capture}
\label{tab:performance}
\begin{adjustbox}{width=\textwidth}
\begin{tabular}{lcccccccc}
\toprule
\textbf{Metric}  &\textbf{Ground Truth}& \textbf{Task} & \textbf{PTM} & \textbf{MAE} & \textbf{Bias} & \textbf{LoA} &  \textbf{TRR} 
&\textbf{ICC$_{2,1}$} \\
\midrule
\multirow{8}{*}{Jump height (cm)}  &\multirow{8}{6em}{Force plates}& \multirow{2}{*}{CMJBL} & $PTM_g$ & 4.87& 1.76& [-10, 13]&  0.95& 0.73\\
  &&  & $PTM_h$ & 2.00& 0.56& [-4.2, 5.4]&  0.95& \textbf{0.96}\\
  \cmidrule{3-9}
  && \multirow{2}{*}{CMJUL} & $PTM_g$ & 3.41& 0.31& [-9.5, 10]&  0.83& 0.50\\
  &&  & $PTM_h$ & 2.29& -0.17& [-6.8, 6.4]&  0.85& \textit{0.75}\\
 \cmidrule{3-9}
  && \multirow{2}{*}{DJBL} & $PTM_g$ & 6.86& 4.99& [-7.7, 18]&  0.87&0.48\\
  &&  & $PTM_h$ & 4.15 & 3.31 & [-3.9, 11] &  0.80 & \textit{0.76} \\
 \cmidrule{3-9}
  && \multirow{2}{*}{DJUL} & $PTM_g$ & 2.89& 1.36& [-5.6, 8.3]&  0.65& 0.25\\
  &&  & $PTM_h$ & 1.71 & 0.36 & [-4.0, 4.7] &  0.39 & \textit{0.69} \\
 \midrule
\multirow{4}{*}{Peak velocity (m/s)}  &\multirow{4}{6em}{OMC}& \multirow{2}{*}{OHP} & $PTM_h$ & 0.14 & -0.05 & [-0.53, 0.42] &  0.80 & 0.87 \\
  &&  & $PTM_b$ & 0.12 & $\approx$0 & [-0.47, 0.47] &  0.81 & \textit{0.87}\\
 \cmidrule{3-9}
  && \multirow{2}{*}{BSQ} & $PTM_h$ & 0.07 & -0.06 & [-0.27, 0.15] &  0.74 & 0.87 \\
  &&  & $PTM_b$ & 0.06& -0.03& [-0.23, 0.16]&  0.75 & \textbf{0.90}\\
 \midrule
\multirow{4}{*}{Mean velocity (m/s)}  &\multirow{4}{6em}{OMC}& \multirow{2}{*}{OHP} & $PTM_h$ & 0.05 & -0.03 & [-0.14, 0.08] &  0.67 & 0.95 \\
  &&  & $PTM_b$ & 0.04 & -0.01 & [-0.11, 0.10] &  0.69 & \textbf{0.97} \\
 \cmidrule{3-9}
  && \multirow{2}{*}{BSQ} & $PTM_h$ & 0.04 & -0.01 & [-0.13, 0.12] &  0.76 & 0.80 \\
  &&  & $PTM_b$ & 0.04 & $\approx$0& [-0.12, 0.13]&  0.77 & \textit{0.80} \\
 \midrule
\multirow{3}{*}{Flight time (secs)}  &\multirow{3}{6em}{Force plates}& DJBL & N/A & 0.11 & 0.11 & [0.06, 0.16] &  0.86 & 0.40 \\
  && DJUL & N/A & 0.09 & 0.09 & [0.05, 0.14] &  0.80 & 0.28 \\
  && RJT & N/A & 0.07& -0.06& [-0.12, 0.005]&  0.52& \textit{0.60}\\
 \cmidrule{2-9}
\multirow{3}{*}{Contact time (secs)}  &\multirow{3}{6em}{Force plates}& DJBL & N/A & 0.12 & -0.12 & [-0.18, -0.05] &  0.53 & 0.37 \\
  && DJUL & N/A & 0.11 & -0.11 & [-0.16, -0.06] &  0.86 & 0.54 \\
  && RJT & N/A & 0.11 & 0.11 & [0.02, 0.19] &  0.90& \textbf{0.80} \\
 \midrule
\multirow{5}{*}{ROM (deg)}  &\multirow{5}{6em}{OMC}& HIR & N/A & 7.84 & -7.60 & [-23, 8] &  0.81 & 0.32 \\
  && HER & N/A & 6.47 & 5.90 & [-3.6, 15] &  0.92 & 0.67 \\
  && ND & N/A & 8.51 & -6.83 & [-22, 7.9] &  0.54 & 0.65 \\
  && SLR & N/A & 6.90 & -5.07 & [-20, 10] &  0.93 & \textbf{0.69}\\
  && SLS & N/A & 43.20 & -37.73& [-1200, 46] &  0.85& 0.27 \\
 \cmidrule{3-9}
\multirow{2}{*}{Angular velocity (deg/s)} &\multirow{2}{6em}{OMC} & ND & N/A & 3.88 & -3.34 & [-9.2, 2.6] &  0.53 & \textbf{0.59} \\
&& SLS & N/A & 60.39 & -50.06 & [-1700, 72] &  0.86 & 0.25 \\
\bottomrule
\end{tabular}
\end{adjustbox}
PTM$_x$: pixel-to-metric conversion using $x$ as reference ($g$: gravity, $h$: participant's height, $b$: barbell length); ROM: range of motion; LoA: limits of agreement; TRR: test-retest reliability. The best ICC for each metric is shown in \textbf{bold} font face, while the best ICC for each task is shown in \textit{italics}.
\end{table*}

The pose data was used to quantify \textbf{four} main types of metrics: \textit{jump height}, which required pixel-to-metric conversion using gravity (PTM$_g$), body height (PTM$_h$), or barbell length (PTM$_b$); \textit{linear velocity}, including mean velocity and peak velocity of tracked parts; \textit{temporal metrics}, including flight time and contact time; and \textit{angular metrics}, including range of motion and angular velocity.

For each task, all repetitions from all participants were taken as individual measurements. For quantitative comparison, we used the \emph{two-way random effects, absolute agreement, single measurement} intraclass correlation coefficient \cite{icc} (ICC$_{2,1}$) and Bland-Altman analysis \cite{bland1986statistical} (BA). For the ICC$_{2,1}$ $\in [0,1]$, values less than 0.5 indicate poor agreement; values between 0.5 and 0.75 indicate moderate agreement; values between 0.75 and 0.9 indicate good agreement; and values greater than 0.90 indicate excellent agreement~\cite{koo_li_2016}.

\begin{figure}
    \centering
    \includegraphics[width=1\linewidth]{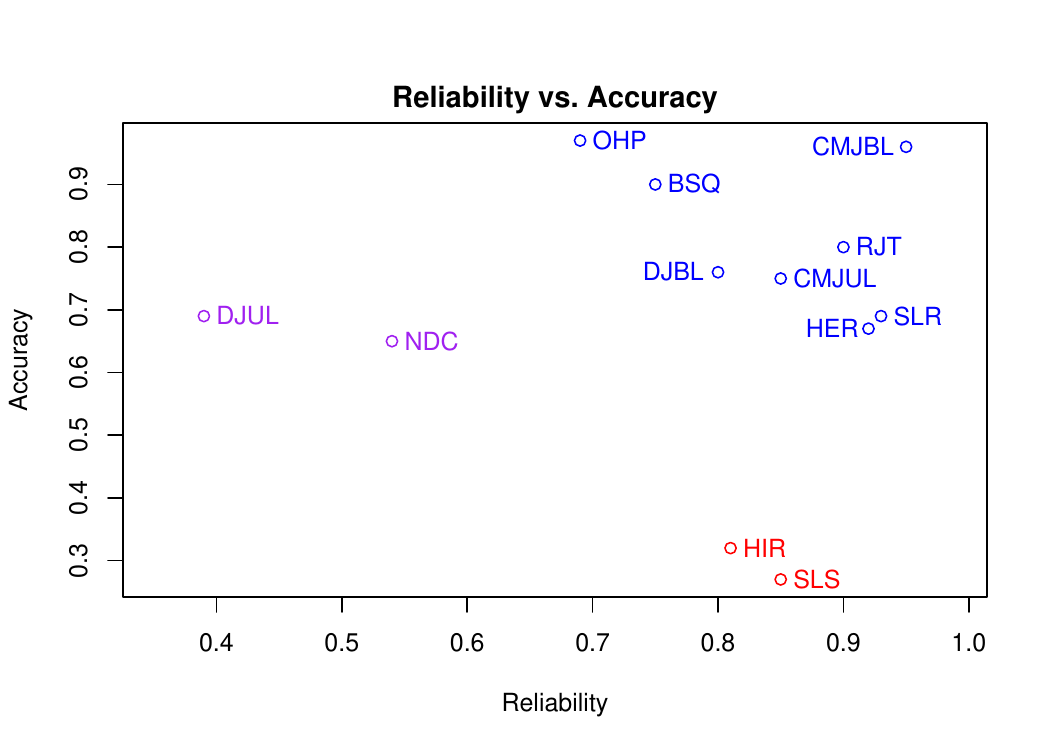}
    \caption{Maximum test-retest reliability versus  ICC (agreement with ground truth)  achieved for each task.}
    \label{fig:rel_vs_acc}
\end{figure}

In a scatterplot of best ICCs against reliability (Fig. \ref{fig:rel_vs_acc}), the tasks form three clusters: \textit{high-accuracy-high-reliability} (HAHR), \textit{low-accuracy-high-reliability} (LAHR), and \textit{low-accuracy-low-reliability} (LALR). For the HAHR tasks, MMC could be used instead of OMC. The two LALR tasks, DJUL and NDC, involved only one limb and were captured from the side view. This often resulted in false limb detections by OpenPose, negatively impacting the reliability and accuracy. During the LAHR tasks (HIR and SLS), MMC showed very low agreement with OMC for the range of motion measurements, although the measurements are highly repeatable. Details of the performance of MMC are given in Table \ref{tab:performance} and discussed in subsequent subsections.

\begin{figure*}[ht]
    \centering
    \begin{subfigure}{0.47\linewidth}
      \includegraphics[width=1\linewidth]{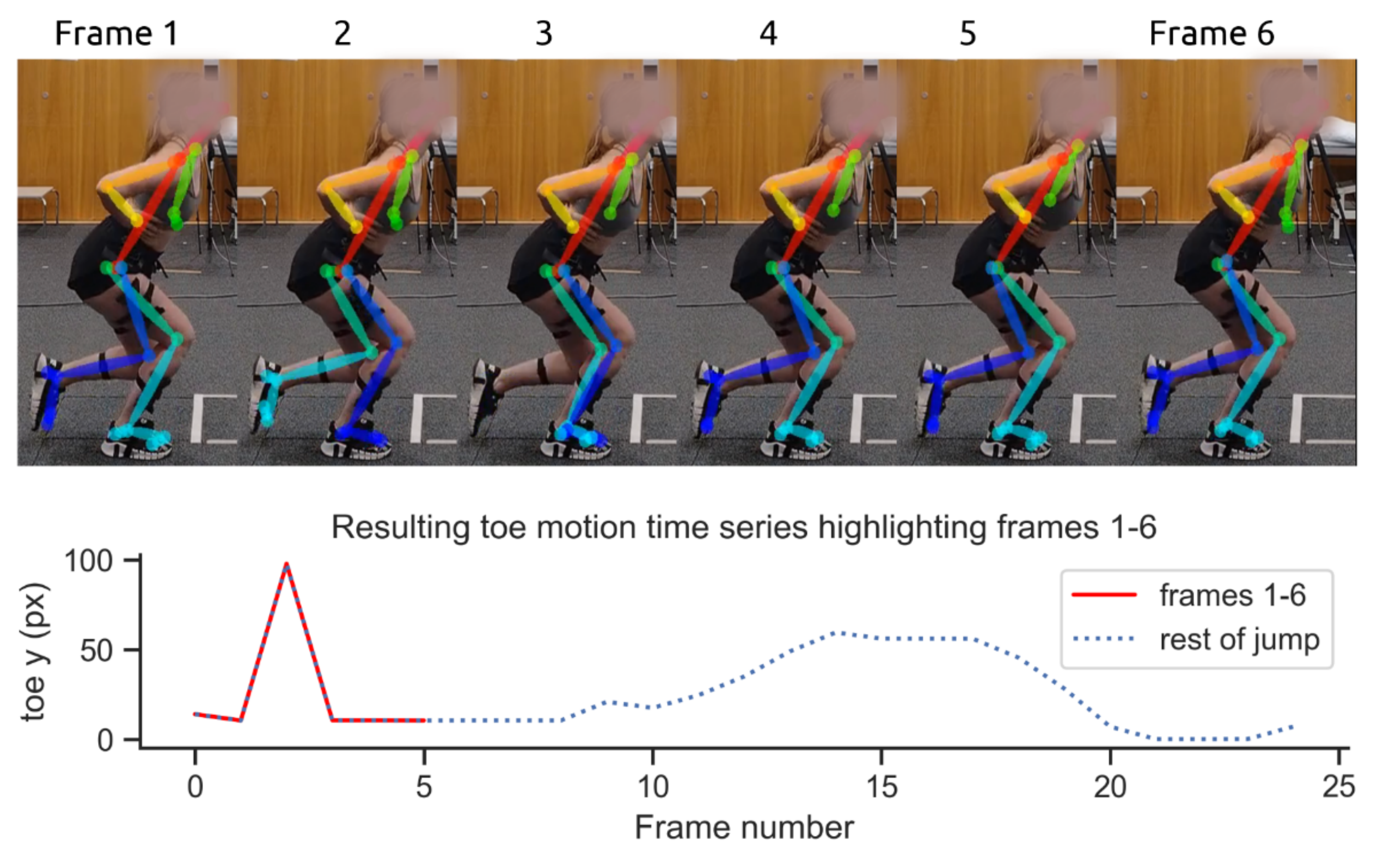}
        \caption{}
    \end{subfigure}
    \begin{subfigure}{0.48\linewidth}
      \includegraphics[width=1\linewidth]{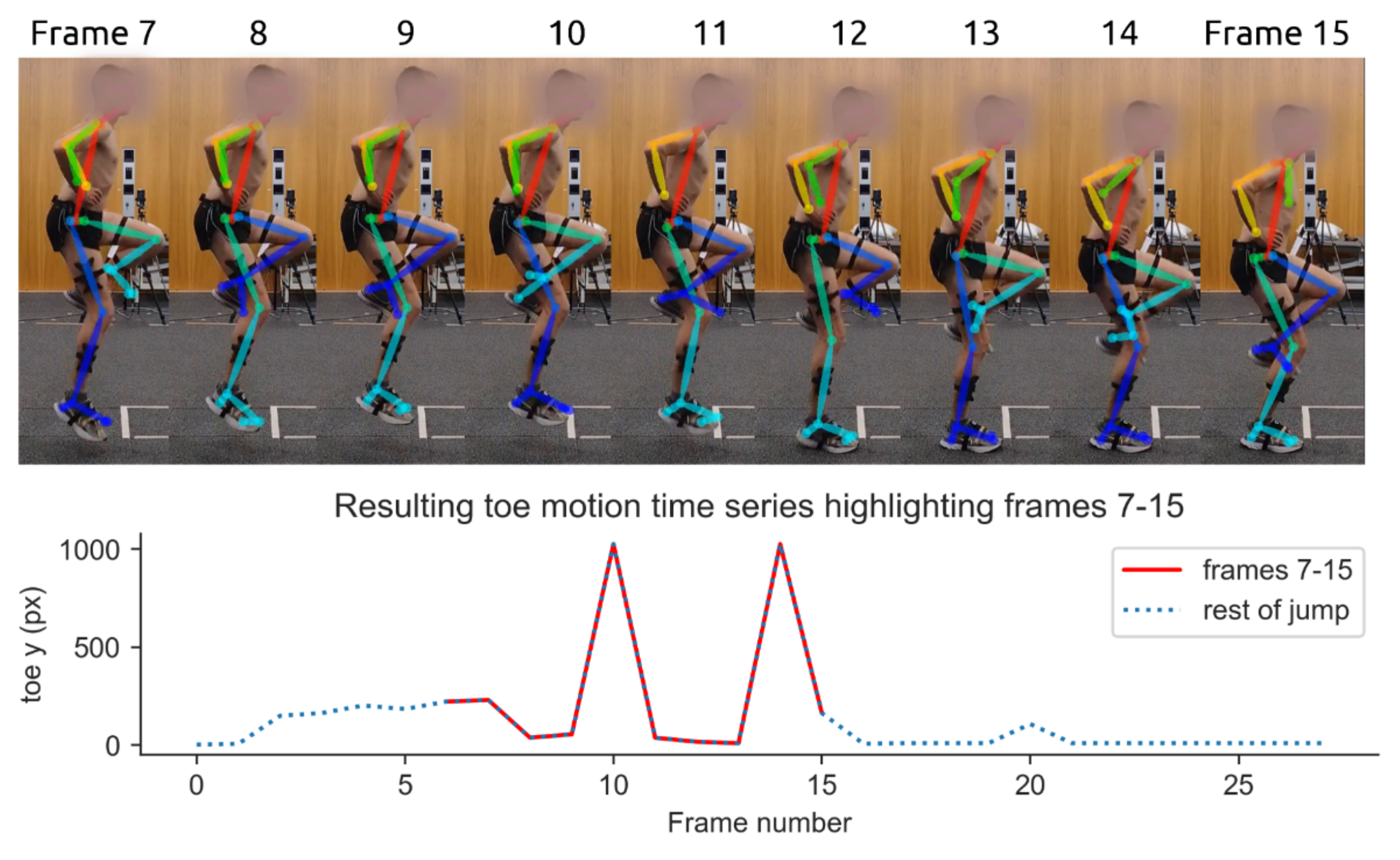}
        \caption{}
    \end{subfigure}
    \caption{Noise in unilateral jumps during pose estimation as seen by observing the limb heatmap colors. (a) In frame 2, the left and right limbs are swapped. In frame 3, the right limb is wrongly detected as two limbs. (b) A failure case showing movements that are not characteristic of countermovement jumps.}
    \label{fig:noisy}
\end{figure*}

As shown in Table \ref{tab:performance}, MMC showed excellent agreement with the force plate in quantifying jump height during the bilateral countermovement jumps, while showing moderate to good agreement during the unilateral countermovement jumps and drop jumps. However, MMC showed poor agreement in measuring unilateral drop jumps. Using body height for pixel-to-millimetre conversion (PTM$_h$) gave better results, and the test-retest reliability (TRR) was higher for the bilateral tasks. The best performance for countermovement and drop jumps was achieved in the front view because the pose estimation model confuses the left and right limbs (Fig. \ref{fig:noisy}) in the side view.

Moreover, MMC also showed good to excellent agreement with optical motion capture during the velocity-based tests. This is because the pose estimation model tracks upper-body parts with higher precision than the lower-body parts. Using the barbell as a reference object (PTM$_b$) showed the best performance in pixel-to-millimetre conversion (Table \ref{tab:performance}).

While MMC showed good promise in quantifying jump height and velocity, its performance in quantifying temporal and angular metrics was not as convincing. For example, although MMC showed good test-retest reliability in quantifying temporal metrics, it showed poor agreement with the force plate in general, except in measuring contact time during the repeated jump test.

Similarly, MMC showed only moderate agreement with OMC in quantifying the range of motion during the seated hip rotation tasks. The limits of agreement (LoA) (Table \ref{tab:performance}) suggest that MMC can easily underestimate or overestimate the ROM measurements. This is a result of the additive effect of errors in each axis. MMC also showed poor agreement with OMC during the single leg squat due to the combined effects of the view, unusual pose, and the high precision requirement for measuring the range of motion. 

\section{Discussion}\label{sec:discussion}
Markerless motion capture showed very good agreement with optical motion capture in measuring velocity and displacement during the overhead press and back squat tasks. In general, the pose estimation model shows very high tracking accuracy for upper body tasks, which results in close agreement with optical motion capture. The high accuracy is also influenced by the possibility of using the barbell as a reference to scale from pixel distances to millimetres. MMC also showed very good agreement with the force plate in measuring vertical jump height during the bilateral countermovement jump. However, MMC showed only moderate agreement with OMC for all the range of motion tasks such as straight leg raise, hip rotation, and Nordic curl. Apparently, greater agreement in tracking angular displacement would require near-perfect tracking in all axes of the pose estimation model. We believe this can only be achieved by custom pose models trained on exercise videos.

The overall accuracy of all the proposed techniques ultimately depends on the accuracy of the pose model used. Pose estimation output is often noisy due to different factors. Some of the main factors identified are video quality, the amount of clutter in the background, video viewpoint, noise in the pose estimation output due to model false positives, and approximations during smoothing, segmentation, and scaling. In particular, pose estimation tends to be especially noisy for tasks with unusual poses such as unilateral countermovement jump and straight leg raise, and tasks that must be analysed on the side view, such as shoulder rotation. In these tasks, pose estimation often confuses the left for the right limbs in the side view. It was also noted that the tracking of lower body keypoints from the knees downwards tends to be noisier than that of upper body keypoints from the hip upwards.

\section{Conclusion}
Based on the results obtained in this study, markerless motion capture with a single smartphone is promising for quantifying simple linear metrics such as vertical jump height and velocity. However, accurately quantifying temporal metrics such as flight time or angular metrics such as range of motion would require much more tracking precision than the current open source pose estimation models offer. Excellent agreement in ROM measurements would require near-perfect agreement in \textbf{all axes pairs} of each capturing device. In addition, movements towards the camera are captured by 3D OMC, but are translated as vertical movements in 2D MMC. The combined effect of errors in individual axes results in larger error orders in degrees. This suggests that 2D MMC may not replace sensors for ROM measurements where high precision is needed, but can instead provide enhanced visual feedback.

\section*{Supplementary information}

Examples of metric measurements are illustrated in Supplementary Figs. 1-8. Participant demographics are summarized in Supplementary Table 1. The code and data necessary to reproduce our analysis, along with additional visualizations, are available on GitHub at \href{https://github.com/mlgig/mvespa}{https://github.com/mlgig/mvespa}.

\bibliography{references}


\section*{Supplementary material}

# Supplementary Materials
# Machine Vision-Enabled Sports Performance Analysis

Timilehin B. Aderinola*[ξ], Hananeh Younesian[ξ], Cathy Goulding, Darragh Whelan,
Brian Caulfield, Georgiana Ifrim

TABLE I: Demography of Participants

| Participant ID | Sex | Mass (kg) | Height (m) | Age (years) |
|---|---|---|---|---|
| IPP01 | Male | 74 | 1.84 | 24 |
| IPP02 | Female | 59 | 1.50 | 25 |
| IPP03 | Male | 74 | 1.73 | 27 |
| IPP04 | Male | 66 | 1.73 | 29 |
| IPP05 | Female | 53 | 1.64 | 28 |
| IPP06 | Male | 63 | 1.67 | 26 |
| IPP07 | Female | 71 | 1.71 | 31 |
| IPP08 | Female | 74 | 1.65 | 41 |
| IPP09 | Male | 65 | 1.80 | 51 |
| IPP10 | Male | 89 | 1.88 | 30 |
| IPP11 | Female | 54 | 1.62 | 25 |
| IPP12 | Male | 72 | 1.85 | 35 |
| IPP13 | Male | 97 | 1.92 | 35 |
| IPP14 | Female | 64 | 1.78 | 26 |
| IPP15 | Male | 83 | 1.84 | 30 |
| IPP16 | Male | 70 | 1.76 | 32 |

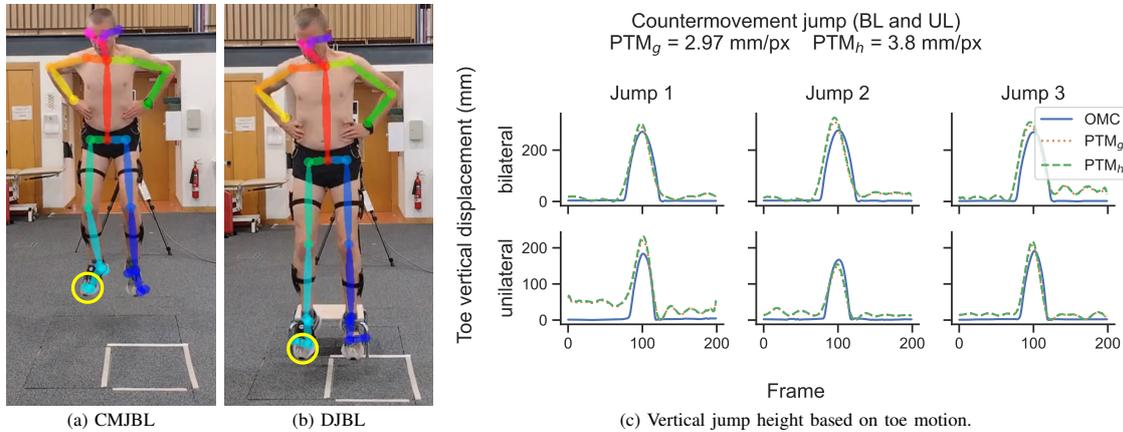

Fig. 1: **Quantifying Jump Height.** Quantifying vertical jump height during countermovement jumps and drop jumps by subtracting the maximum vertical displacement of the toe (circled) from its resting position in the vertical axis.

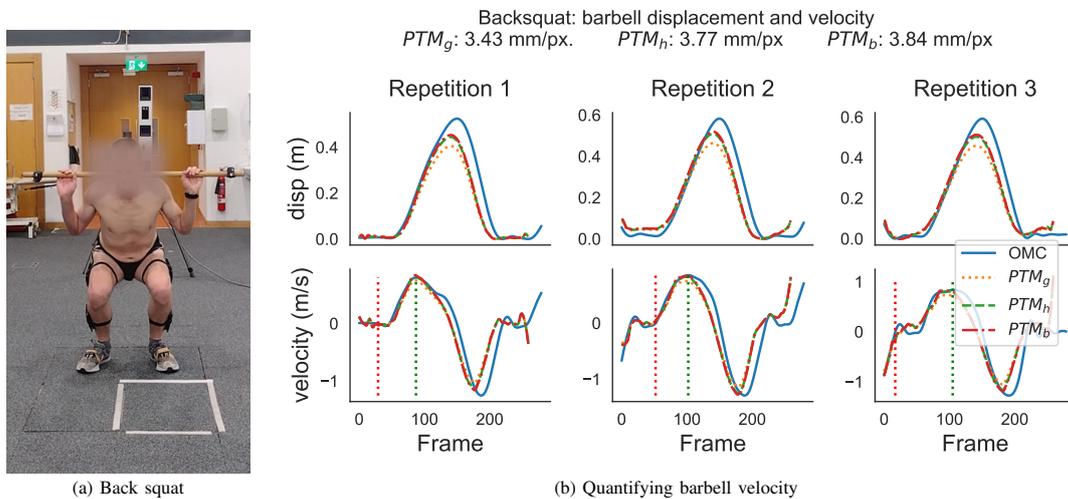

Fig. 2: **Velocity-Based Testing**. Measuring barbell displacement and velocity during velocity-based tests. The concentric phases are bounded using red (start) and green (end) vertical dashes.

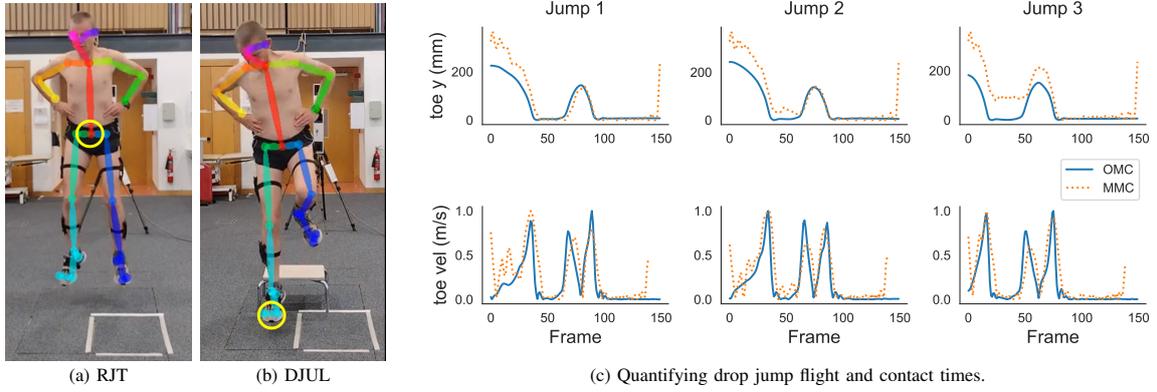

(a) RJT  (b) DJUL  (c) Quantifying drop jump flight and contact times.

Fig. 3: **Quantifying Temporal Metrics**. For the drop jumps, the contact time was obtained as the distance between the first two peaks of the toe velocity, while the flight time is the distance between the last two peaks.

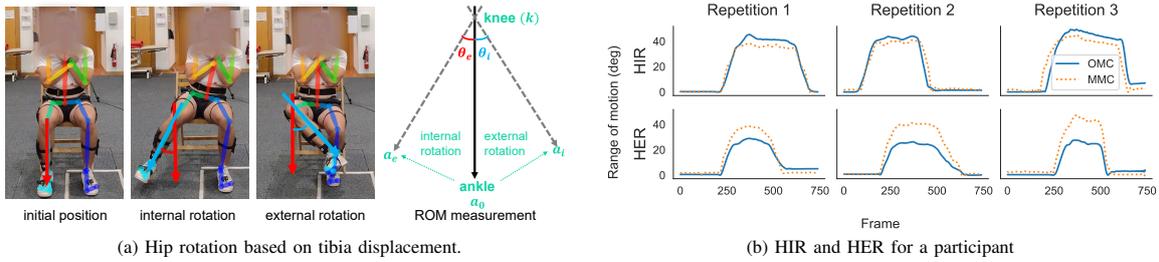

(a) Hip rotation based on tibia displacement.  (b) HIR and HER for a participant

Fig. 4: **Hip Range of Motion**. Measuring seated hip internal (HIR) and external rotation (HER) using the maximum angular displacement at the knee as a proxy.

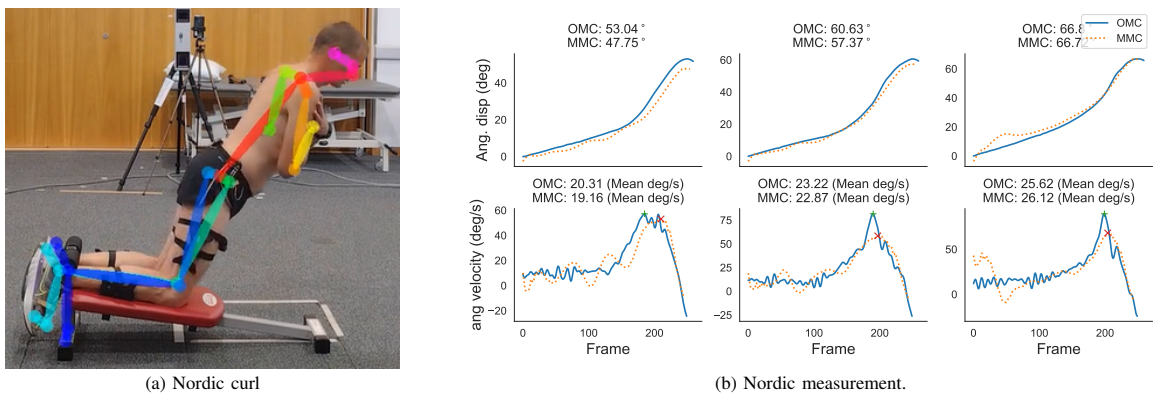

(a) Nordic curl  (b) Nordic measurement.

Fig. 5: **Nordic Curl**. Measuring angular displacement and velocity during the Nordic curl.

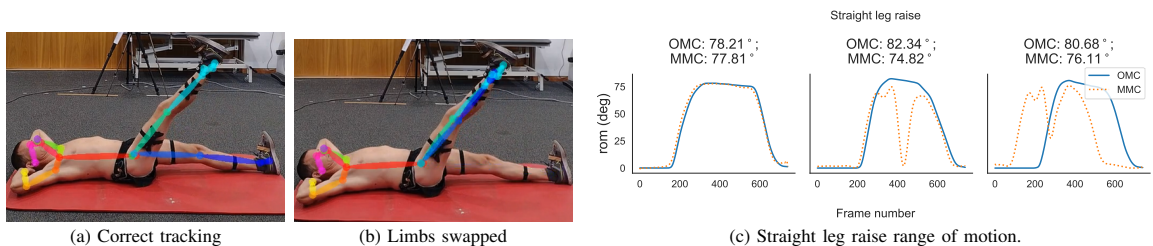

(a) Correct tracking (b) Limbs swapped (c) Straight leg raise range of motion.

Fig. 6: **Straight Leg Raise**. (a) Both limbs are correctly detected and tracked. (b) The right limb is wrongly detected as the left limb while the left limb is undetected.

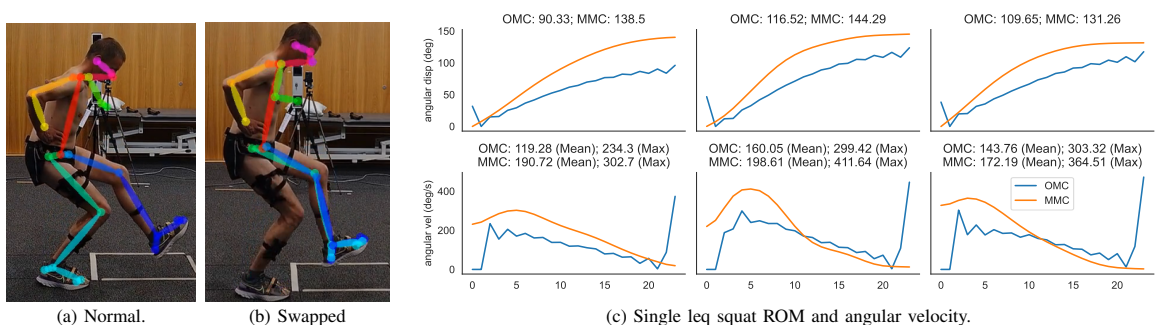

(a) Normal. (b) Swapped (c) Single leg squat ROM and angular velocity.

Fig. 7: **Single Leg Squat**. An example of single leg squat shown for a participant showing instances where the limbs are correctly tracked, and where the left and right limbs are confused.

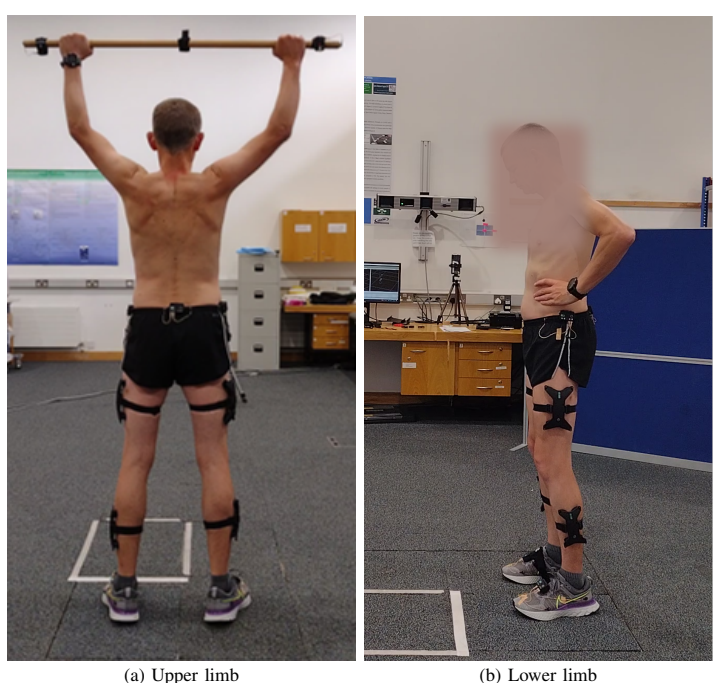

(a) Upper limb (b) Lower limb

Fig. 8: **Marker and Cluster Placement**. Codamotion clusters and marker placement.



\end{document}